\documentclass[letterpaper]{article} 

\usepackage{aaai24}
\usepackage{times}  
\usepackage{helvet}  
\usepackage{courier}  
\usepackage[hyphens]{url}  
\usepackage{graphicx} 
\usepackage{natbib}  
\usepackage{caption} 
\usepackage[noend]{algpseudocode}
\usepackage{algorithmicx,algorithm}
\usepackage{amsmath,amssymb,amssymb}
\usepackage{booktabs} 
\usepackage{mathrsfs}
\usepackage{multirow}
\usepackage{comment}
\usepackage{changepage}
\usepackage{colortbl} 
\usepackage{arydshln} 
\usepackage{makecell}
\usepackage{newfloat}
\usepackage{listings}
\DeclareCaptionStyle{ruled}{labelfont=normalfont,labelsep=colon,strut=off} 
\floatstyle{ruled}
\newfloat{listing}{tb}{lst}{}
\floatname{listing}{Listing}
\title{DualTeacher: Bridging Coexistence of Unlabelled Classes for Semi-supervised Incremental Object Detection}

\author{
    Ziqi Yuan\textsuperscript{\rm 1}\thanks{Equal contribution.},
    Liyuan Wang\textsuperscript{\textrm{1}${\ast}$},
    Wenbo Ding\textsuperscript{\rm 2},
    Xingxing Zhang\textsuperscript{\rm 1},
    Jiachen Zhong\textsuperscript{\rm 2}
    Jianyong Ai\textsuperscript{\rm 2},
    Jianmin Li\textsuperscript{\textrm{1}\textrm{\dag}},
    Jun Zhu\textsuperscript{\textrm{1}}\thanks{Corresponding authors.}
}

\affiliations{
    \textsuperscript{\rm 1}Dept. of Comp. Sci. \& Tech., Institute for AI, BNRist Center, THBI Lab, Tsinghua-Bosch Joint Center for ML, Tsinghua University, Beijing, China.
    \textsuperscript{\rm 2}SAIC AI Lab, Shanghai, China.\\
     yzq22@mails.tsinghua.edu.cn, wly19@tsinghua.org.cn, lijianmin@mail.tsinghua.edu.cn,\\
     dingwenbowoai@gmail.com, xxzhang2020@mail.tsinghua.edu.cn, julightzhong10@outlook.com,\\
     514424280@qq.com, dcszj@mail.tsinghua.edu.cn
}

\usepackage{bibentry}

\usepackage[switch]{lineno}

\begin{document}

\maketitle


\begin{abstract}
In real-world applications, an object detector often encounters object instances from new classes and needs to accommodate them effectively. Previous work formulated this critical problem as incremental object detection (IOD), which assumes the object instances of new classes to be fully annotated in incremental data. However, as supervisory signals are usually rare and expensive, the supervised IOD may not be practical for implementation. In this work, we consider a more realistic setting named semi-supervised IOD (SSIOD), where the object detector needs to learn new classes incrementally from a few labelled data and massive unlabelled data without catastrophic forgetting of old classes. A commonly-used strategy for supervised IOD is to encourage the current model (as a student) to mimic the behavior of the old model (as a teacher), but it generally fails in SSIOD because a dominant number of object instances from old and new classes are coexisting and unlabelled, with the teacher only recognizing a fraction of them. Observing that learning only the classes of interest tends to preclude detection of other classes, we propose to bridge the coexistence of unlabelled classes by constructing two teacher models respectively for old and new classes, and using the concatenation of their predictions to instruct the student. This approach is referred to as DualTeacher, which can serve as a strong baseline for SSIOD with limited resource overhead and no extra hyperparameters. We build various benchmarks for SSIOD and perform extensive experiments to demonstrate the superiority of our approach (e.g., the performance lead is up to 18.28 AP on MS-COCO). Our code is available at \url{https://github.com/chuxiuhong/DualTeacher}.
\end{abstract}

\section{Introduction}
The ability of incremental learning is critical for deep neural networks to accommodate real-world dynamics, but is limited by catastrophic forgetting of old knowledge \cite{mcclelland1995there,wang2023comprehensive}. 
Especially in object detection tasks, there is usually a large number of object instances from new classes that need to be incorporated and recognized. Numerous efforts have been devoted to the setting of incremental object detection (IOD) \cite{shmelkov2017incremental_ilod,peng2020faster,li2019rilod,feng2022overcoming}, where the object detector attempts to learn new classes from large amounts of labelled data on the basis of remembering old classes. A commonly-used strategy is to perform \emph{knowledge distillation} from a frozen copy of the old model: The current model acts as a student to learn the new classes, while the old model acts as a teacher (denoted as the \emph{old teacher}) to stabilize the predictions of old classes. 
Since the object instances of new classes have been fully annotated, the student can faithfully inherit the predictions of old classes and thus achieves satisfactory performance.

However, the incremental data are often partially-labelled in many real-world applications. For example, self-driving cars collect massive amounts of data every day, and it is expensive and impractical to annotate them completely. Household robots need to accommodate customized scenarios through limited instructions from the user, while frequent requests for labelling may affect the user experience. 
In response to the practical challenge of labelling scarcity, we consider here a more realistic setting where new classes are annotated for only a small fraction of incremental data, referred to as semi-supervised IOD (SSIOD).

\begin{figure}[t]
    \centering
    \vspace{+0.2cm}
    \includegraphics[width=0.93\linewidth]{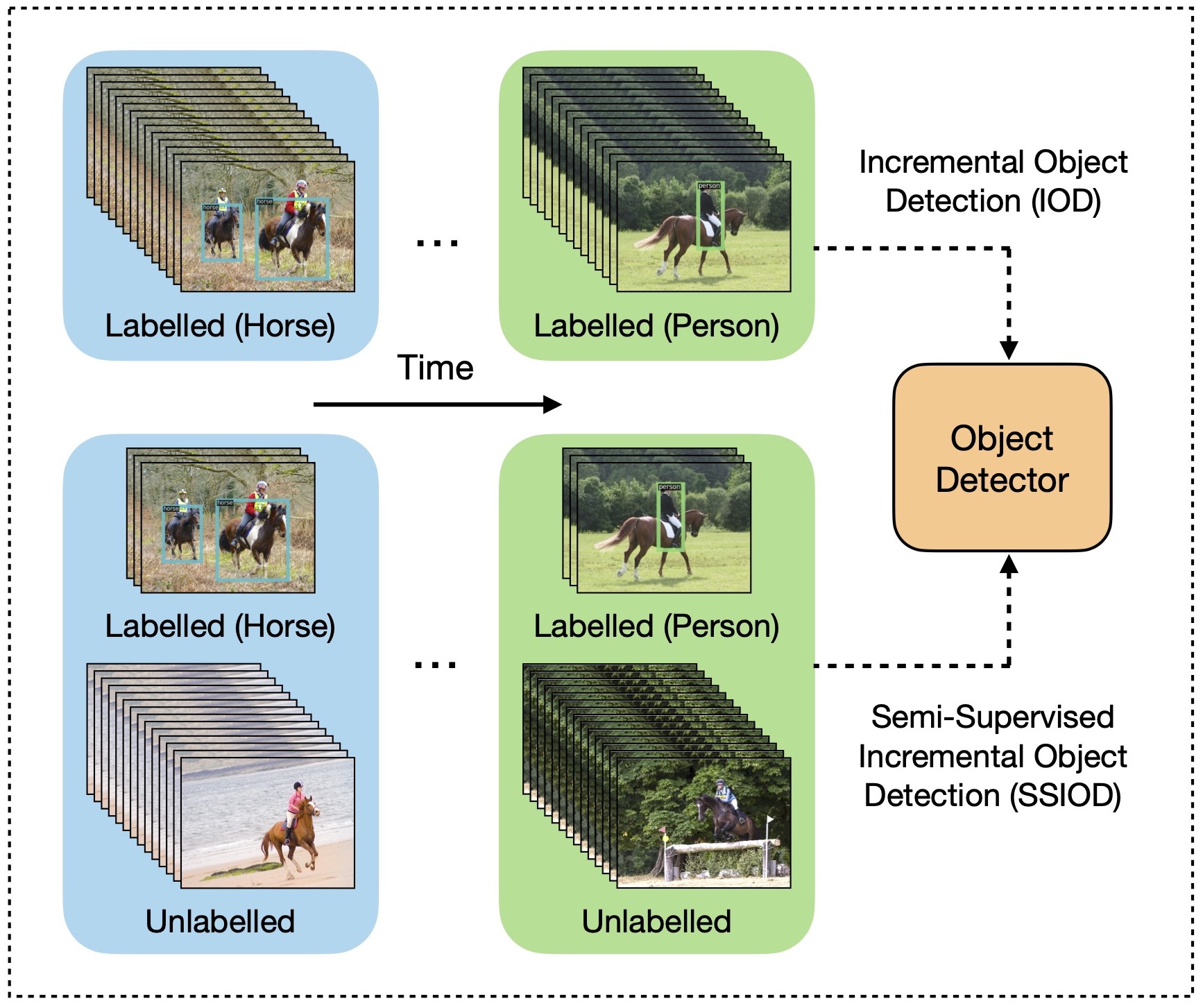}
\caption{Comparison of IOD and SSIOD.} 
\vspace{-0.5cm}
\label{IOD_SSIOD}
\end{figure}

\begin{figure*}[th]
    \centering
    \includegraphics[width=0.95\linewidth]{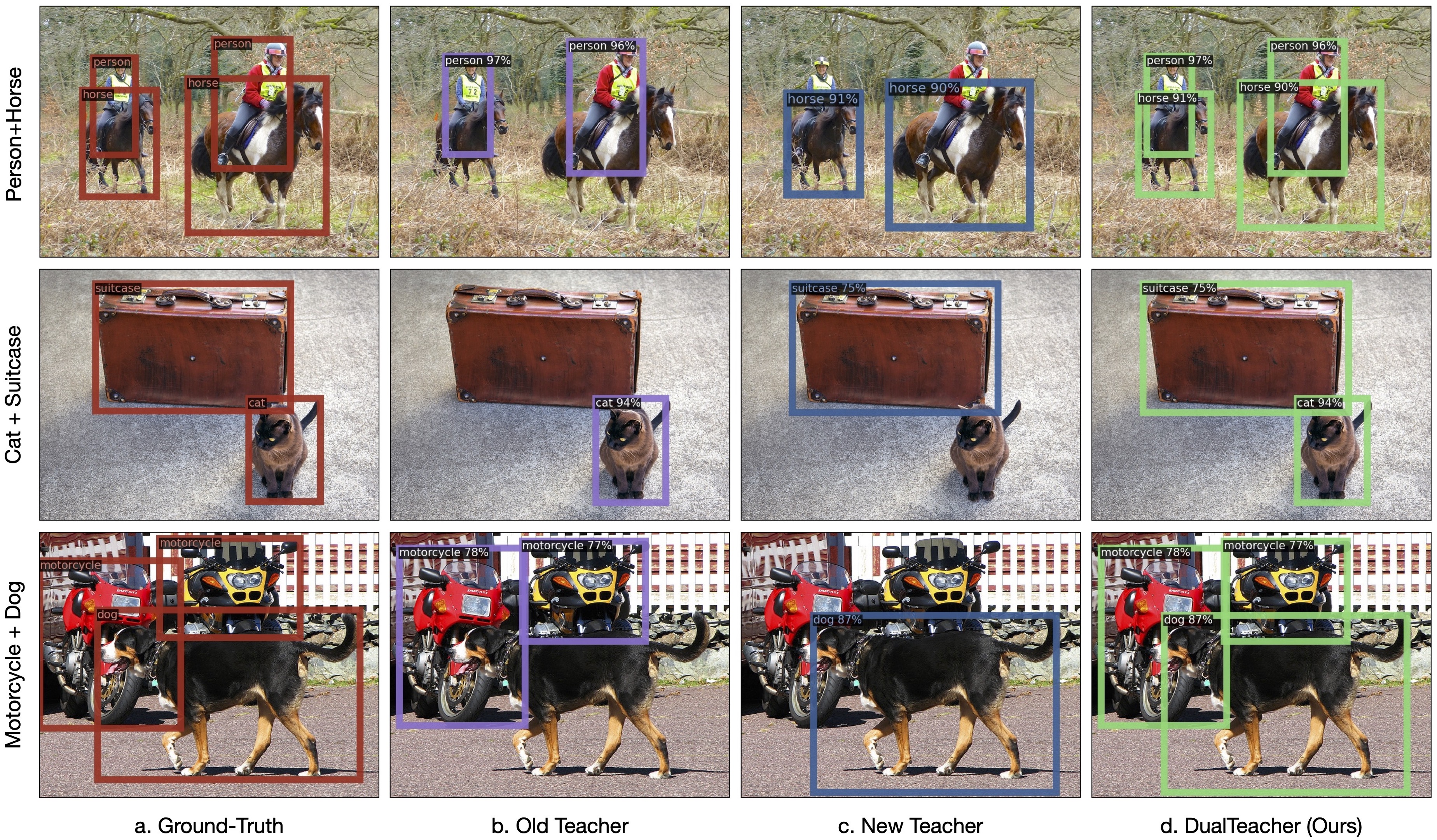}
\caption{Comparison of object detection results for different teacher models.} 
\vspace{-0.2cm}
\label{detection_result}
\end{figure*}

In contrast to supervised IOD, SSIOD can access to limited supervision of new classes and massive unlabelled data (see Fig.~\ref{IOD_SSIOD}). It is well known that unlabelled data is crucial for semi-supervised learning (SSL) but difficult to exploit. Current SSL techniques for object detection \cite{liu2021unbiased,xu2021end,sohn2020simple,liu2022unbiased,wang2023consistent} generally exploit labelled data to train a teacher model (denoted as the \emph{new teacher}), the teacher model provides pseudo-labels of unlabelled data to train a student model, and the student model progressively updates the teacher model with exponential moving average (EMA). 
In SSIOD, the new teacher suffers from catastrophic forgetting as incremental labelled data have only supervision of new classes. Nevertheless, implementing the old teacher can only marginally improve or even deteriorate the performance of SSIOD, since the object instances of old and new classes usually coexist in \emph{incremental unlabelled data} while the two teachers with limited and disjointed knowledge can only recognize some of them (see Fig.~\ref{detection_result}, b, c).


The disjoint knowledge stems from the non-overlapping annotations in incremental labelled data, i.e., only the currently learned classes are annotated and the other classes are left as the ``background''. As a result, the old teacher has learned to both identify old classes and ``ignore'' new classes, and vice versa for the new teacher. This property leads to their conflicting predictions for incremental unlabelled data, but also prevents low-quality predictions of uncertain classes. In order to bridge the coexistence of unlabelled classes, here we propose DualTeacher as a simple but effective approach for SSIOD.
Specifically, we construct the old teacher that can identify old classes and train a new teacher with labelled data of new classes. Then we use the concatenation of their predictions as pseudo-labels (see Fig.~\ref{detection_result}, d) to train a student model with unlabelled data, and progressively update the new teacher with EMA.
Therefore, the new teacher can obtain knowledge of both old and new classes from incremental unlabelled data and further improve its predictions to benefit the student model.
We perform extensive experiments to demonstrate the superiority of DualTeacher in SSIOD, which clearly outperforms strong IOD baselines under different labelling ratios and task splits (e.g., the improvement is up to \textbf{18.28} AP on MS-COCO). 

Our contributions include: (1) To the best of our knowledge, we are the \emph{first} to consider semi-supervised incremental object detection (SSIOD), which is practical for updating object detectors in realistic applications; (2) We attribute the central challenge of SSIOD to the conflict in predicting old and new classes that coexist in incremental unlabelled data, and propose a simple but effective approach to address it; (3) We build a variety of SSIOD benchmarks and extensively validate the superiority of our approach.

\section{Related Work}

\textbf{Incremental Learning}, also called continual learning or lifelong learning, aims to acquire, accumulate and exploit knowledge from sequentially arrived tasks \cite{wang2023comprehensive,parisi2019continual,wang2023incorporating}. Previous work has focused on supervised settings of image classification tasks, overcoming catastrophic forgetting of old tasks when learning each new task. Representative strategies include selectively stabilization of parameter changes \cite{kirkpatrick2017overcoming,wang2021afec}, replay of representative old training samples \cite{rebuffi2017icarl,wang2021memory}, construction of task-specific parameters \cite{serra2018overcoming,wang2022coscl,wang2023hierarchical}, etc. Real-world applications typically face more complex tasks, as well as the expense and scarcity of supervisory signals. 
To cope with limited supervision of incremental data, semi-supervised continual learning (SSCL) has been considered for image classification tasks \cite{wang2021ordisco}. In SSCL, many representative continual learning strategies become less effective in exploiting incremental unlabelled data, indicating the non-trivialness of this particular challenge. An alternative approach is to recover old data distributions by training an auxiliary generative model \cite{wang2021ordisco}, but requires significant resource overheads.

\textbf{Object Detection} is one of the most important tasks in computer vision and has received widespread attention \cite{lin2017feature,he2017mask}. This is often more complex and realistic than image classification, where multiple object instances appearing in one image need to be correctly located and identified. Mainstream object detectors can be conceptually categorized into one-stage detectors, e.g., YOLO \cite{redmon2016you}, and two-stage detectors, e.g., Faster R-CNN \cite{ren2015faster}, depending on whether localization and classification are performed separately. IOD \cite{shmelkov2017incremental_ilod} is a typical setting of incremental learning for object detection, where the model learns object instances of new classes in a supervised manner.
Since the new classes are fully annotated in incremental data, most IOD methods \cite{shmelkov2017incremental_ilod,peng2020faster,li2019rilod,feng2022overcoming} performed knowledge distillation from a frozen copy of the old model to stabilize the predictions of the old classes and thus overcome catastrophic forgetting.

\textbf{Semi-supervised Learning} aims to learn effectively from partially-labelled data, which is typically achieved by input augmentation and consistency regularization \cite{berthelot2019mixmatch,laine2016temporal,tarvainen2017mean}. Compared to image classification, semi-supervised learning techniques for object detection are often limited by the complexity of architecture design in object detectors \cite{liu2021unbiased}. Most existing methods for semi-supervised object detection are implemented with a two-stage object detector (especially Faster R-CNN), using a teacher-student framework with interdependent updates \cite{liu2021unbiased,xu2021end,sohn2020simple,liu2022unbiased,wang2023consistent}. Despite practical significance, the extension of incremental learning to semi-supervised object detection remains to be explored. The most relevant setting is open-world object detection (OWOD) \cite{joseph2021towards}, which first identifies unknown classes and then incorporates their labels. However, OWOD needs to annotate all object instances of unknown classes, which may not be practical for massive unlabelled data emerging in real-world applications.

\section{Semi-supervised Incremental Object Detection}
In this section, we introduce the problem formulation of SSIOD and necessary preliminaries.

\subsection{Problem Formulation}
In contrast to supervised IOD, semi-supervised IOD (SSIOD) annotates the object instances of new classes for only a small fraction of training samples and leaves the rest unlabelled, where the object instances of old and new classes may coexist in each data point (see Fig.~\ref{IOD_SSIOD}).
Let's denote $\mathcal{D}_t = \mathcal{D}_t^l \cup \mathcal{D}_t^u$ as the training dataset of task $t$, which is also called an ``incremental phase'', provided sequentially to an object detector $f$ parameterized by $\theta$. The labelled subset $\mathcal{D}_t^l = \{(x_{t,n}^l, y_{t,n}^l)\}_{n=1}^{N_t^l}$ have $N_t^l$ data-label pairs, where the annotation $y_{t,n}^l$ of a labelled image $x_{t,n}^l$ includes the locations, sizes and classes of object instances (i.e., all bounding boxes). The unlabelled subset $\mathcal{D}_t^u = \{(x_{t,n}^u)\}_{n=1}^{N_t^u}$ have $N_t^u$ unlabelled data. 
Here we further denote $\mathcal{C}_t^l$ as all labelled classes of $y_{t,n}^l$, and assume that the old and new classes are disjoint, i.e., $\mathcal{C}_i^l \cap \mathcal{C}_j^l = \emptyset$ for $i \neq j$.\footnote{This assumption is inherited from supervised IOD, and can be further relaxed to non-overlapping annotations of classes in general, i.e., $\mathcal{C}_i^l \neq \mathcal{C}_j^l$ for $i \neq j$.}
The goal of the object detector is to correctly predict the location and identity of the object instances belonging to all classes ever seen on the test set.

\subsection{Preliminaries of Object Detection}

\textbf{Two-Stage Detector:} 
Without loss of generality, here we focus on the two-stage R-CNN detector, which is widely-used for implementing semi-supervised learning techniques for object detection \cite{liu2021unbiased,xu2021end,sohn2020simple,liu2022unbiased,wang2023consistent}. Specifically, the object detector $f$ can be represented as a composition of functions $(f_{{\rm{roi}}} \circ f_{{\rm{rpn}}} \circ f_{{\rm{b}}})$, parameterized by $\theta_{\rm{roi}}$, $\theta_{\rm{rpn}}$ and $\theta_{\rm{b}}$, respectively. The backbone $f_{{\rm{b}}}$ projects an input image $x_{t,n}$ to its feature map $h_{t,n} \in \mathbb{R}^{C \times H \times W}$, where $C$, $H$ and $W$ denote the number of channels, height and width, respectively. 
The region proposal network (RPN) $f_{{\rm{rpn}}}$ proposes regions of potential object instances from the feature map, specifying objectiveness scores and bounding box locations for multiple proposals. The RoI head $f_{{\rm{roi}}}$ predicts the classes for these proposals and regresses their bounding box locations as the final output.

\begin{figure*}[th]
    \centering
    \includegraphics[width=0.90\linewidth]{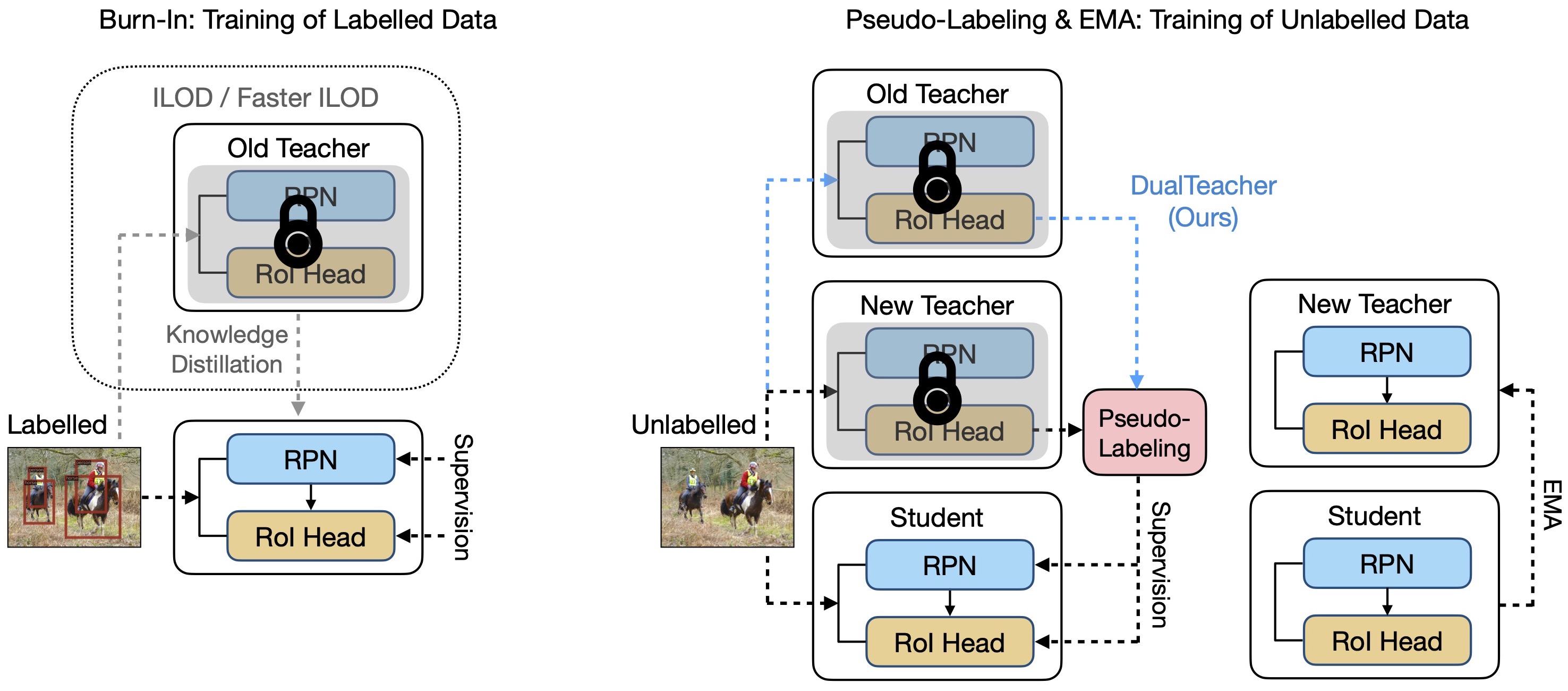}
\caption{Comparison of representative methods for IOD (gray dashed arrow) and our DualTeacher for SSIOD (blue dashed arrow). The black dashed arrow denotes the general framework of semi-supervised learning for object detection. The locked and shaded areas indicate that the parameters are fixed.} 
\vspace{-0.2cm}
\label{dualteacher}
\end{figure*}

\textbf{Semi-supervised Learning:}
In order to effectively leverage partially-labelled data in $\mathcal{D}_t$, mainstream semi-supervised learning techniques construct two interdependent models as a teacher-student framework, denoted as $f_{{\rm{tea}}}$, $\theta_{{\rm{tea}}}$ and $f_{{\rm{stu}}}$, $\theta_{{\rm{stu}}}$, respectively.
Specifically, the object detector $f$ first learns labelled data $\mathcal{D}_t^l$ through the RPN classification loss $\mathcal{L}_{{\rm{rpn}}}^{{\rm{cls}}}$, the RPN regression loss $\mathcal{L}_{{\rm{rpn}}}^{{\rm{reg}}}$, the ROI classification loss $\mathcal{L}_{{\rm{roi}}}^{{\rm{cls}}}$ and the ROI regression loss $\mathcal{L}_{{\rm{roi}}}^{{\rm{reg}}}$ \cite{ren2015faster}, serving as a \emph{burn-in} stage:
\begin{equation}
\begin{split}
    \mathcal{L}_{{\rm{sup}}} = \frac{1}{N_t^l} & \sum_{n=1}^{N_t^l} [\mathcal{L}_{{\rm{rpn}}}^{{\rm{cls}}}(x_{t,n}^l, y_{t,n}^l) + \mathcal{L}_{{\rm{rpn}}}^{{\rm{reg}}}(x_{t,n}^l, y_{t,n}^l) \\
    &+ \mathcal{L}_{{\rm{roi}}}^{{\rm{cls}}}(x_{t,n}^l, y_{t,n}^l) + \mathcal{L}_{{\rm{roi}}}^{{\rm{reg}}}(x_{t,n}^l, y_{t,n}^l)].
\label{eq_sup_loss}
\end{split}
\end{equation}
In practice, the standard cross-entropy for multi-class classification can be replaced by Focal loss \cite{lin2017focal} to alleviate class imbalance \cite{liu2021unbiased}. Next, $f$ initializes $f_{{\rm{tea}}}$ and $f_{{\rm{stu}}}$ with the obtained weights $\theta_{{\rm{tea}}} \leftarrow \theta$ and $\theta_{{\rm{stu}}} \leftarrow \theta$, respectively. The initialized $f_{{\rm{tea}}}$ generates pseudo-labels for each $x_{t,n}^u$ in $\mathcal{D}_t^u$:
\begin{equation}
\hat{y}_{t,n}^u = f_{{\rm{tea}}}(x_{t,n}^u).
\end{equation}
At this point, $\theta_{{\rm{tea}}}$ in $f_{{\rm{tea}}}$ is fixed and only $\theta_{{\rm{stu}}}$ in $f_{{\rm{stu}}}$ is updated via backward propagation:
\begin{equation}
    \theta_{{\rm{stu}}} \leftarrow \theta_{{\rm{stu}}} + \alpha_{{\rm{stu}}} \frac{\partial ( \mathcal{L}_{{\rm{sup}}}+\lambda_{{\rm{unsup}}} \mathcal{L}_{{\rm{unsup}}})}{\partial \theta_{{\rm{stu}}}},
\label{eq_stu_update}
\end{equation}
\begin{equation}
    \mathcal{L}_{{\rm{unsup}}} = \frac{1}{N_t^u} \sum_{n=1}^{N_t^u} [\mathcal{L}_{{\rm{rpn}}}^{{\rm{cls}}}(x_{t,n}^u, \hat{y}_{t,n}^u) + \mathcal{L}_{{\rm{roi}}}^{{\rm{cls}}}(x_{t,n}^u, \hat{y}_{t,n}^u)],
    \label{eq_unsup_loss}
\end{equation}
where $\lambda_{{\rm{unsup}}}$ is a hyperparameter to balance the learning of labelled and unlabelled data, and $\alpha_{{\rm{stu}}}$ is the learning rate of $\theta_{{\rm{stu}}}$.
Every time $f_{{\rm{stu}}}$ is updated over several iterations, $f_{{\rm{tea}}}$ is refined via exponential moving average (EMA): \footnote{Recent work in semi-supervised learning for object detection has improved the implementation of EMA \cite{liu2021unbiased,liu2022unbiased}. Here we present only its original form.}
\begin{equation}
    \theta_{{\rm{tea}}} \leftarrow \alpha_{{\rm{tea}}} \theta_{{\rm{tea}}} + (1-\alpha_{{\rm{tea}}})\theta_{{\rm{stu}}},
    \label{eq_tea_ema}
\end{equation}
where $\alpha_{{\rm{tea}}}$ is a hyperparameter to determine the magnitude of updates.
Therefore, the quality of pseudo-labels is also progressively improved from unlabelled data.

\textbf{Incremental Learning:} 
With only the current training dataset available, the object detector tends to catastrophically forget the old knowledge. To overcome this issue, a commonly-used strategy is to create a frozen copy of the old model, denoted as $f_{{\rm{old}}}$, and add a knowledge distillation loss to penalize differences in output between $f$ and $f_{{\rm{old}}}$.\footnote{For clarity, we refer to $f_{{\rm{new}}}$ and $f_{{\rm{old}}}$ specifically as the \emph{new teacher} and the \emph{old teacher} in the context of SSIOD, respectively.} The knowledge distillation loss can generally be defined as
\begin{equation}
\mathcal{L}_{{\rm{sup}}}^{{\rm{kd}}} = \frac{1}{N_t^l} \sum_{n=1}^{N_t^l} d[f(x_{t,n}^l), f_{{\rm{old}}}(x_{t,n}^l)],
\label{eq_sup_kd}
\end{equation}
where $d$ is the function of measuring differences, and the supervised loss $\mathcal{L}_{{\rm{sup}}}$ becomes 
\begin{equation}
\mathcal{L}_{{\rm{sup}}}' = \mathcal{L}_{{\rm{sup}}}+ \lambda_{{\rm{sup}}}^{{\rm{kd}}} \mathcal{L}_{{\rm{sup}}}^{{\rm{kd}}}, 
\label{eq_sup_sup_kd}
\end{equation}
with a hyperparameter $\lambda_{{\rm{sup}}}^{{\rm{kd}}}$ to control the strength of $\mathcal{L}_{{\rm{sup}}}^{{\rm{kd}}}$.
As an initial attempt, ILOD \cite{shmelkov2017incremental_ilod} performed knowledge distillation of $f_{{\rm{roi}}}$ (i.e., the classification and regression outputs), quantified by the $L_2$ distance. In response to the architecture of Faster R-CNN, Faster ILOD \cite{peng2020faster} further performed knowledge distillation of $f_{{\rm{b}}}$ and $f_{{\rm{rpn}}}$, quantified by the $L_1$ and $L_2$ distances, respectively. 
There are many other methods to design the knowledge distillation loss depending on the architecture and output of different object detectors \cite{peng2021sid,li2019rilod,feng2022overcoming,zhang2020class}.
Since the object instances of old and new classes usually coexist in incremental data, $\mathcal{L}_{{\rm{sup}}}^{{\rm{kd}}}$ and $\mathcal{L}_{{\rm{sup}}}$ allow for an explicit balance between them.

For SSIOD, the knowledge distillation loss can be naturally adapted to the unlabelled data in each $\mathcal{D}_t^u$:
\begin{equation}
\mathcal{L}_{{\rm{unsup}}}^{{\rm{kd}}} = \frac{1}{N_t^u} \sum_{n=1}^{N_t^u} d[f_{{\rm{stu}}}(x_{t,n}^u), f_{{\rm{old}}}(x_{t,n}^u)].
\label{eq_unsup_kd}
\end{equation}
Correspondingly, the unsupervised loss $\mathcal{L}_{{\rm{unsup}}}$ becomes 
\begin{equation}
\mathcal{L}_{{\rm{unsup}}}' = \mathcal{L}_{{\rm{unsup}}}+ \lambda_{{\rm{unsup}}}^{{\rm{kd}}} \mathcal{L}_{{\rm{unsup}}}^{{\rm{kd}}}, 
\label{eq_unsup_unsup_kd}
\end{equation}
which balances the old and new classes with a hyperparameter $\lambda_{{\rm{unsup}}}^{{\rm{kd}}}$ similarly to $\mathcal{L}_{{\rm{sup}}}'$. The optimization of the student model $f_{{\rm{stu}}}$ in Eq.~\ref{eq_stu_update} becomes
\begin{equation}
    \theta_{{\rm{stu}}} \leftarrow \theta_{{\rm{stu}}} + \alpha_{{\rm{stu}}} \frac{\partial ( \mathcal{L}_{{\rm{sup}}}'+\lambda_{{\rm{unsup}}} \mathcal{L}_{{\rm{unsup}}}')}{\partial \theta_{{\rm{stu}}}}.
\label{eq_stu_update_iod}
\end{equation}
In fact, $\mathcal{L}_{{\rm{sup}}}^{{\rm{kd}}}$ in Eq.~\ref{eq_sup_kd}~/~\ref{eq_sup_sup_kd} and $\mathcal{L}_{{\rm{unsup}}}^{{\rm{kd}}}$ in Eq.~\ref{eq_unsup_kd}~/~\ref{eq_unsup_unsup_kd} are computationally and functionally similar in updating $f_{{\rm{stu}}}$ with Eq.~\ref{eq_stu_update_iod}, since $x_{t,n}^l$ and $x_{t,n}^u$ follow the same distribution. Therefore, the implementation of knowledge distillation (KD) to SSIOD can be conceptually separated as three ways: using $\mathcal{L}_{{\rm{sup}}}^{{\rm{kd}}}$ for $f_{{\rm{tea}}}$ ($f_{{\rm{tea}}}$+KD), using $\mathcal{L}_{{\rm{sup}}}^{{\rm{kd}}}$ and $\mathcal{L}_{{\rm{unsup}}}^{{\rm{kd}}}$ for $f_{{\rm{stu}}}$ ($f_{{\rm{stu}}}$+KD), and using both (Both+KD).


\section{Method}

\begin{figure}[t]
    \centering
    \includegraphics[width=0.98\linewidth]{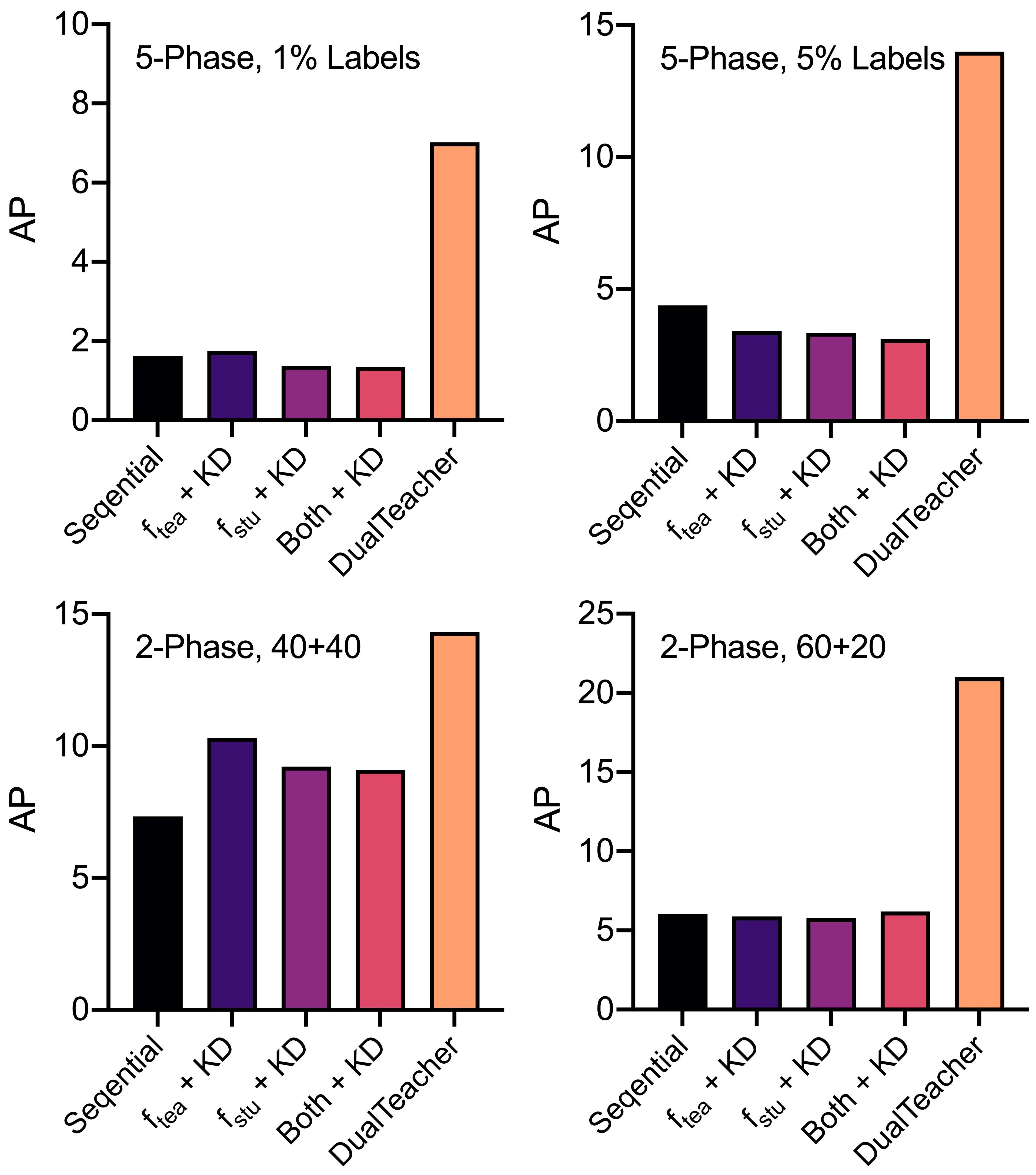}
    \vspace{-0.1cm}
\caption{Performance of SSIOD. Here we take Faster ILOD as the baseline approach of knowledge distillation (KD), which can be implemented to $f_{{\rm{tea}}}$, $f_{{\rm{stu}}}$, or both.} 
\vspace{-0.4cm}
\label{detection_ap}
\end{figure}

Despite performing well in supervised IOD, the commonly-used strategy of knowledge distillation (KD) fails in SSIOD. Here we conduct experiments on the MS-COCO dataset \cite{lin2014microsoft} with different labelling ratios and task splits. 
As shown in Fig.~\ref{detection_ap}, the use of KD in all three possible ways only marginally improves or even deteriorates the performance, compared to sequential training that suffers from severe catastrophic forgetting. In particular, KD tends to perform worse when handling a larger number of incremental phases, further limiting its capability and flexibility to accommodate real-world changes. In response to the specific challenges of SSIOD, we provide a more in-depth analysis and present our approach as below.

Because of the limited supervision, the object detector must learn each task effectively from massive unlabelled data. However, the use of either $\mathcal{L}_{{\rm{sup}}}^{{\rm{kd}}}$ or $\mathcal{L}_{{\rm{unsup}}}^{{\rm{kd}}}$ conflicts with this objective. 
Since the old teacher $f_{{\rm{old}}}$ has been optimized for only recognizing old classes, it cannot make reliable predictions for object instances of new classes. Therefore, when updating the student $f_{{\rm{stu}}}$ with Eq.~\ref{eq_stu_update_iod}, $\mathcal{L}_{{\rm{sup}}}^{{\rm{kd}}}$ and $\mathcal{L}_{{\rm{unsup}}}^{{\rm{kd}}}$ for remembering the old classes would severely compromise $\mathcal{L}_{{\rm{sup}}}$ and $\mathcal{L}_{{\rm{unsup}}}$ for learning the new classes, which can be seen as a particular dilemma between \emph{stability} and \emph{plasticity}.
In addition, since the unsupervised loss $\mathcal{L}_{{\rm{unsup}}}$ for updating $f_{{\rm{stu}}}$ is calculated from the pseudo-labels of $f_{{\rm{tea}}}$ (Eq.~\ref{eq_unsup_loss}) and $f_{{\rm{tea}}}$ is progressively refined from $f_{{\rm{stu}}}$ (Eq.~\ref{eq_tea_ema}), the errors arising from $\mathcal{L}_{{\rm{sup}}}^{{\rm{kd}}}$ and $\mathcal{L}_{{\rm{unsup}}}^{{\rm{kd}}}$ would \emph{accumulate} in semi-supervised learning.

\begin{algorithm}[tb]
    \caption{Training Algorithm of DualTeacher}
    \label{alg:algorithm}
    \textbf{Input}: Object detector $f$ with parameters $\theta$, training datasets $\mathcal{D}_1, ..., \mathcal{D}_t$, number of tasks $T$, number of epochs $e_1$ and $e_2$, hyperparameters $\alpha_{{\rm{tea}}}$, $\alpha_{{\rm{stu}}}$ and $\lambda_{{\rm{unsup}}}$ for semi-supervised learning. \\ 
    \textbf{Output}: Parameters $\theta_{{\rm{tea}}}$ and $\theta_{{\rm{stu}}}$
    \begin{algorithmic}[1] 
          \For{$t = 1, ..., T$}
              \For{$epoch = 1, ..., e_1$} 
              \State Optimize $\theta$ in $f$ with Eq.~\ref{eq_sup_loss} 
              \EndFor
              \If{$t = 1$}
              \State Initialize $\theta_{{\rm{tea}}} \leftarrow \theta$ and $\theta_{{\rm{stu}}} \leftarrow \theta$
              \Else
              \State  Initialize $\theta_{{\rm{tea}}} \leftarrow \theta$
              \EndIf
              \For{$epoch = 1, ..., e_2$} 
              \State Generate pseudo-labels with Eq.~\ref{eq_pseudo_dt}
              \State Optimize the student $\theta_{{\rm{stu}}}$ with Eq.~\ref{eq_stu_dt} 
              \State Update the new teacher $\theta_{{\rm{tea}}}$ with Eq.~\ref{eq_tea_ema}
              \EndFor
              \State Update the old teacher $\theta_{{\rm{old}}} \leftarrow \theta_{{\rm{tea}}}$
        \EndFor
       \State \textbf{return} $(\theta_{{\rm{tea}}}, \theta_{{\rm{stu}}})$
    \end{algorithmic}
\end{algorithm}

The particular dilemma between stability and plasticity stems from the \emph{non-overlapping annotations} in incremental labelled data $\mathcal{D}_t^l$, and is exposed in processing incremental unlabelled data $\mathcal{D}_t^u$. Although the object instances of old and new classes usually coexist, only the currently learned classes are annotated and the other classes are left as the ``background''. As a result, the old teacher $f_{{\rm{old}}}$ is optimized for recognizing old classes and ignoring potential new classes, while the new teacher $f_{{\rm{tea}}}$ is optimized for recognizing new classes and ignoring potential old classes (see Fig.~\ref{detection_result}, b, c). This property causes their predictions to conflict with each other when updating the student $f_{{\rm{stu}}}$ with unlabelled data, but also prevents low-quality predictions of uncertain classes. 
A desirable model should recognize the object instances of all classes ever seen, which in this case corresponds to the concatenation of predictions specific to each task or incremental phase (see Fig.~\ref{detection_result}, a, d).

Motivated by this, we propose a simple but effective approach to bridge coexistence of unlabelled classes in SSIOD (detailed in Algorithm~1). 
Specifically, we treat the old teacher $f_{{\rm{old}}}$ and the new teacher $f_{{\rm{tea}}}$ as a couple of DualTeacher, which specialize in recognizing the object instances of old and new classes, respectively.
Then we use the concatenation of their predictions as pseudo-labels for unlabelled data to update the student $\theta_{{\rm{stu}}}$:
\begin{equation}
\hat{y}_{t,n}^{{\rm{dt}}}=f_{{\rm{old}}}(x_{t,n}^u) \cup f_{{\rm{tea}}}(x_{t,n}^u).
\label{eq_pseudo_dt}
\end{equation}
The unsupervised loss $\mathcal{L}_{{\rm{unsup}}}$ becomes
\begin{equation}
    \mathcal{L}_{{\rm{unsup}}}^{{\rm{dt}}} = \frac{1}{N_t^u} \sum_{n=1}^{N_t^u} [\mathcal{L}_{{\rm{rpn}}}^{{\rm{cls}}}(x_{t,n}^u, \hat{y}_{t,n}^{{\rm{dt}}}) + \mathcal{L}_{{\rm{roi}}}^{{\rm{cls}}}(x_{t,n}^u, \hat{y}_{t,n}^{{\rm{dt}}})],
    \label{eq_unsup_loss_dt}
\end{equation}
and the optimization of $\theta_{{\rm{stu}}}$ becomes
\begin{equation}
    \theta_{{\rm{stu}}} \leftarrow \theta_{{\rm{stu}}} + \alpha_{{\rm{stu}}} \frac{\partial ( \mathcal{L}_{{\rm{sup}}}+\lambda_{{\rm{unsup}}} \mathcal{L}_{{\rm{unsup}}}^{{\rm{dt}}} ) }{\partial \theta_{{\rm{stu}}}}.
    \label{eq_stu_dt}
\end{equation}
Both $f_{{\rm{old}}}$ and $f_{{\rm{tea}}}$ are frozen when updating $f_{{\rm{stu}}}$, while $f_{{\rm{tea}}}$ is then updated by EMA as in Eq.~\ref{eq_tea_ema}. Therefore, $f_{{\rm{stu}}}$ obtains correct and compatible supervision when processing unlabelled data, so as to improve its capability. $f_{{\rm{tea}}}$ also progressively obtains the knowledge of recognizing old and new classes, where the former is consistent with $f_{{\rm{old}}}$ and will not affect the concatenation. Note that our DualTeacher does \emph{not} introduce additional hyperparameters, which is essential for implementation since the regular hyperparameter search may not be available in incremental learning \cite{chaudhry2018efficient_agem}. 
Besides, our DualTeacher can serve as a plug-in module for representative semi-supervised learning techniques\footnote{Some methods may differ slightly in the form of loss functions.} in SSIOD, thanks to the generality of the teacher-student framework (see Fig.~\ref{dualteacher}).



\newcommand{\tabincell}[2]{\begin{tabular}{@{}#1@{}}#2\end{tabular}}

\begin{table}[t]
	\centering
    \caption{The performance of 5-Phase SSIOD with different labelling ratios. The number of new classes is balanced by task. ``-R'' stands for replay of all labelled data.} 
      \vspace{-0.1cm}
	\smallskip
      \renewcommand\arraystretch{1.7}
	\resizebox{0.47\textwidth}{!}{ 
	\begin{tabular}{c|l|cccccc}
	 \hline
        Setting& Baseline &${\rm{AP}}$ &${\rm{AP}}_{\rm{50}}$ &${\rm{AP}}_{{\rm{75}}}$ & ${\rm{AP}}_S$ & ${\rm{AP}}_M$ & ${\rm{AP}}_L$ \\
        \hline
       \multirow{6}*{\tabincell{c}{1\% \\ Labels}} 
        &Sequential & 1.62 & 2.92 & 1.69 & 0.82 & 1.95 & 1.70 \\
        &ILOD & 1.42 & 2.61 & 1.42 & 0.58 & 1.95 & 1.28 \\
        &Faster ILOD & 1.34 & 2.34 & 1.45 & 0.54 & 1.55 & 1.87\\
        &DualTeacher & \textbf{10.65} & \textbf{20.81} & \textbf{9.69} & \textbf{4.70} & \textbf{11.95} & \textbf{13.29} \\
        \cdashline{2-8}[2pt/2pt]
        &Sequential-R & 10.20 & 18.04 & 10.04 &4.67 &10.71  &12.69  \\
        &DualTeacher-R & \textbf{11.41} & \textbf{20.95} & \textbf{11.08} & \textbf{5.42} & \textbf{11.90} & \textbf{14.93}  \\
       \hline
        \multirow{6}*{\tabincell{c}{5\% \\ Labels}} 
        &Sequential & 4.38 & 7.11 & 4.67 & 1.91 & 4.99 & 5.75 \\
        &ILOD & 3.18 & 5.58 & 3.36 & 1.37 & 3.81 & 4.02 \\
        &Faster ILOD & 3.10 & 5.43 & 3.14 & 1.37 & 3.53 & 4.05 \\
        &DualTeacher & \textbf{16.99} & \textbf{30.60} & \textbf{17.07} & \textbf{7.34} & \textbf{18.51} & \textbf{22.62} \\
        \cdashline{2-8}[2pt/2pt]
        &Sequential-R & 18.88 & 32.78 & 19.33 & 9.07& 19.71 & 24.47 \\
        &DualTeacher-R & \textbf{19.93} & \textbf{34.71} &\textbf{20.58} & \textbf{10.20} & \textbf{20.82} & \textbf{26.14} \\
       \hline
	\end{tabular}
	} 
	\label{table_5_phase}
	\vspace{-0.1cm}
\end{table}

\section{Experiment}
In this section, we first describe experimental setups for SSIOD, and then present the experimental results.

\textbf{Benchmark:} We construct a variety of SSIOD benchmarks based on the MS-COCO dataset \cite{lin2014microsoft}, which is commonly used for both semi-supervised learning and incremental learning. MS-COCO includes object instances from 80 classes with 118287 images for training and 5000 images for testing. Following the protocol of supervised IOD \cite{shmelkov2017incremental_ilod,peng2020faster}, we split them into multiple incremental phases, with the training images for each phase containing only the annotations of new classes. For SSIOD, we only use a small fraction of labelled data and leave the rest unlabelled. In other words, only a few images have annotations of new classes rather than old classes, while the others have no annotations (see Fig.~\ref{IOD_SSIOD}). In experiments, we analyze the impact of labelling ratios (e.g., 1\% or 5\%) \cite{liu2021unbiased} and task splits (e.g., 40-40 or 60-20) \cite{feng2022overcoming} under the settings of 5- and 2-phase, respectively.

\textbf{Implementation:} Because a majority of semi-supervised learning (SSL) techniques for object detection \cite{liu2021unbiased,xu2021end,sohn2020simple,liu2022unbiased} are implemented in a Faster R-CNN architecture \cite{ren2015faster}, we focus on a similar architecture in this paper. Here we choose Unbiased Teacher \cite{liu2021unbiased,liu2022unbiased} as the default method for semi-supervised learning, which is representative and easily applicable to other SSL techniques. We adapt the official implementation of Unbiased Teacher to SSIOD. Specifically, we use a ResNet-50 backbone, initialized by the ImageNet pre-trained model, set the confidence threshold $\delta=0.7$, and apply a similar protocol of data augmentation. More training details are provided in Appendix.

\textbf{Evaluation Metric:}
We report the average precision (AP) with a wide range of IoU thresholds as the evaluation metric. Specifically, ${\rm{AP}}$, ${\rm{AP}}_{\rm{50}}$ and ${\rm{AP}}_{{\rm{75}}}$ apply the IoU threshold of 0.50 to 0.95, 0.50 and 0.75, respectively. ${\rm{AP}}_S$, ${\rm{AP}}_M$ and ${\rm{AP}}_L$ calculate ${\rm{AP}}$ for object instances with different pixel areas, corresponding to smaller than $32^2$, $32^2$ to $96^2$ and larger than $96^2$, respectively.

\begin{table}[t]
	\centering
    \caption{The performance of 2-Phase SSIOD with different task splits. The labelling ratio is 5\%. ``-R'' stands for replay of all labelled data.} 
      \vspace{-0.1cm}
	\smallskip
      \renewcommand\arraystretch{1.7}
	\resizebox{0.47\textwidth}{!}{ 
	\begin{tabular}{c|l|cccccc}
	 \hline
        Setting& Baseline & ${\rm{AP}}$ & ${\rm{AP}}_{\rm{50}}$ &${\rm{AP}}_{{\rm{75}}}$ & ${\rm{AP}}_S$ & ${\rm{AP}}_M$ & ${\rm{AP}}_L$ \\
        \hline
       \multirow{6}*{\tabincell{c}{40+40 \\ Classes}} 
        &Sequential & 7.33 & 12.47 & 7.69 & 2.14 & 7.46 & 11.80 \\
        &ILOD & 9.33 & 16.35 & 9.58 & 3.60 & 10.03  & 13.08 \\
        &Faster ILOD & 9.09 & 16.01 & 9.35 & 3.25 & 9.46 & 12.71 \\
        &DualTeacher & \textbf{17.94} & \textbf{31.84} & \textbf{18.21} & \textbf{6.59} & \textbf{19.49} & \textbf{25.89} \\
        \cdashline{2-8}[2pt/2pt]
        &Sequential-R & 17.81 & 31.14 & 18.28 & 6.95 & 19.06 & 25.35  \\
        &DualTeacher-R  & \textbf{17.39} & \textbf{30.42} & \textbf{17.79} & \textbf{7.15} & \textbf{18.31}  & \textbf{25.54} \\
       \hline
        \multirow{6}*{\tabincell{c}{60+20 \\ Classes}} 
        &Sequential & 6.06 & 9.73 & 6.53 & 2.01 & 6.09 & 8.31 \\
        &ILOD & 6.21 & 10.66 & 6.37 & 1.95 & 5.93  & 8.83  \\
        &Faster ILOD & 6.20 & 10.69 & 6.42 & 2.00 & 5.75 & 8.98 \\
        &DualTeacher & \textbf{24.34} & \textbf{42.31} & \textbf{24.95} & \textbf{11.87} & \textbf{26.29} & \textbf{32.11} \\
        \cdashline{2-8}[2pt/2pt]
        &Sequential-R & 23.70 & 40.55 & 24.41 & 11.93 & 25.52  & 32.38 \\
        &DualTeacher-R & \textbf{24.27} & \textbf{42.17} & \textbf{25.06} & \textbf{13.11} & \textbf{26.00} & \textbf{32.15} \\
       \hline
	\end{tabular}
	} 
	\label{table_2_phase}
	\vspace{-0.1cm}
\end{table}

\textbf{Baseline:} 
We first report the performance of sequential training, which represents the lower bound performance of incremental learning with severe catastrophic forgetting. Then we consider two representative IOD methods that can be applied to Faster R-CNN, such as ILOD \cite{shmelkov2017incremental_ilod} and Faster ILOD \cite{peng2020faster}. They both created a frozen copy of the old model to perform knowledge distillation. ILOD regularized the final outputs of classification and regression for two-stage object detectors. Faster~ILOD was designed specifically for Faster R-CNN and additionally regularized the differences in intermediate outputs.
We further evaluate the effect of replaying labelled data, which proved to be an effective strategy for semi-supervised continual learning of image classification tasks \cite{wang2021ordisco}.

\begin{figure}[t]
    \centering
    \includegraphics[width=1\linewidth]{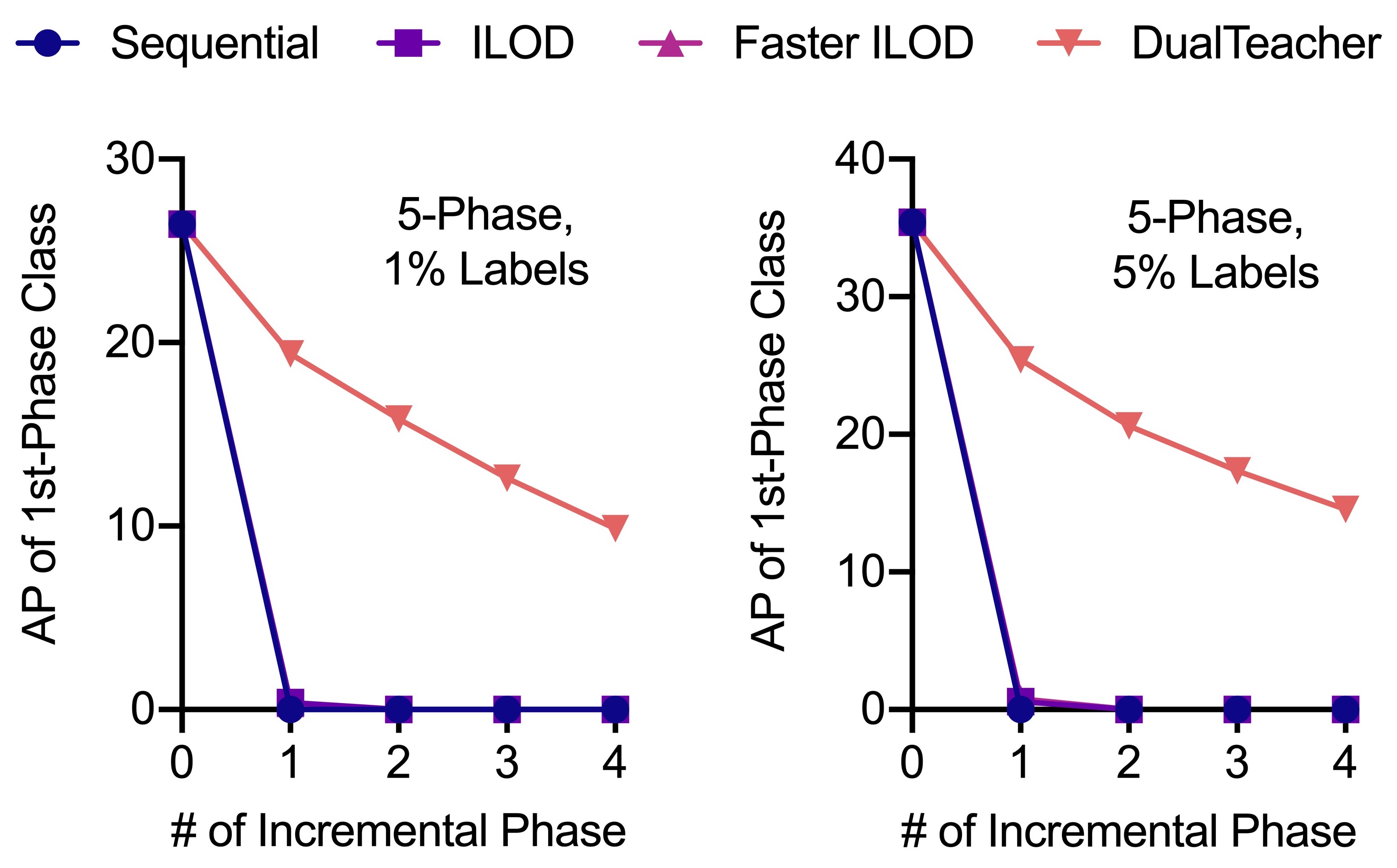}
    \vspace{-0.2cm}
\caption{Subsequent performance of the classes learned in the initial phase. The number of new classes is balanced by task. ``-R'' stands for replay of all labelled data.} 
\vspace{-0.4cm}
\label{forgetting_curve}
\end{figure}

\textbf{Hyperparameter:}
Through an extensive hyperparameter search, we observe that representative IOD methods can hardly improve or even interfere with the performance of SSIOD due to the use of inappropriate knowledge distillation. 
To ensure the reliability of our implementation, we first construct a task sequence with only labelled data, which is identical to supervised IOD \cite{peng2020faster}, and validate that such methods can indeed improve sequential training in the setting intended by their design. 
Then we perform SSIOD experiments with the best hyperparameters selected for all methods.
Note that the commonly-used protocol of selecting hyperparameters requires multiple runs of the task sequence for grid search, which may not be available for incremental learning in realistic scenarios \cite{chaudhry2018efficient_agem}. We still use this protocol for other methods in order to give them the best chance. In contrast, our DualTeacher does not introduce additional hyperparameters on the top of semi-supervised learning, which is clearly more practical and easier to implement.

\textbf{Overall Performance:} As shown in Table~\ref{table_5_phase} and \ref{table_2_phase}, our DualTeacher demonstrates a significant performance lead (up to \textbf{18.28} AP) in both 5-phase and 2-phase settings of SSIOD. Specifically, representative IOD methods (e.g., ILOD and Faster ILOD) only slightly improve or even interfere with the performance of sequential training, where the improvement only occurs in 2-phase setting and becomes much smaller for 40+40 classes. These results suggest that the use of knowledge distillation in SSIOD would have an increasingly negative impact with the addition of new classes. In contrast, our DualTeacher consistently delivers strong performance gains with different labelling ratios and task splits.
We further explore the effect of experience replay. Consistent with the results of image classification \cite{wang2021ordisco}, replay of all labelled data is indeed remarkably effective for continual / incremental learning in semi-supervised scenarios. Since all labelled data have been revisited, the SSIOD problem becomes overcoming catastrophic forgetting of the unlabelled parts of incremental data. On the top of this strategy, our DualTeacher can still improve the performance by a clear margin, validating its motivation. 
Besides, the semi-supervised learning performance (i.e., joint training of partially-labelled data) of Unbiased Teacher \cite{liu2021unbiased} is 20.75 AP and 28.27 AP for 1\% labels and 5\% labels, respectively. We expect subsequent work in SSIOD to further close the performance gap.

\begin{figure}[t]
    \centering
    \includegraphics[width=0.55\linewidth]{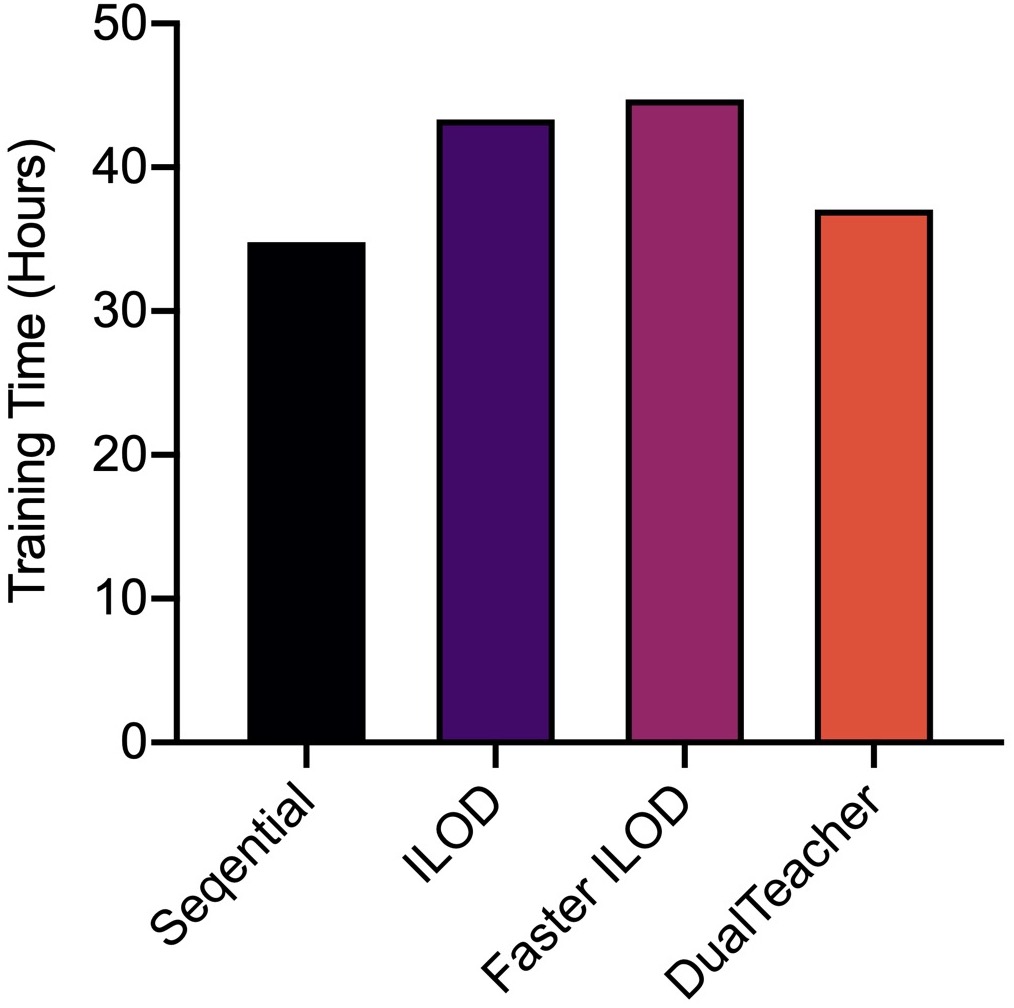}
    \vspace{-0.1cm}
\caption{Comparison of training time on 5-phase SSIOD with 5\% labels, measured by the same four-card Tesla V100.} 
\vspace{-0.2cm}
\label{resource_overhead}
\end{figure}

\textbf{Detailed Analysis:} As our DualTeacher only introduces the old teacher's predictions to generate pseudo-labels, the performance improvement over sequential training has provided an explicit ablation study. Below, we analyze its advantages more extensively. First, we present the subsequent performance of the first-phase classes to evaluate the extent of \emph{catastrophic forgetting}. As shown in Fig.~\ref{forgetting_curve}, sequential training, ILOD and Faster ILOD all suffer from severe catastrophic forgetting, where the performance of old classes drops to almost zero in subsequent learning phases. Our DualTeacher can largely alleviate catastrophic forgetting, with the performance of old classes decaying slowly. Second, we evaluate the \emph{resource overhead} in terms of storage and computation. The storage overhead of all methods is equivalent to that of the object detector itself.
Meanwhile, our DualTeacher creates smaller computational overheads than ILOD and Faster ILOD (see Fig.~\ref{resource_overhead}) but achieves significantly better performance in SSIOD. Together with the advantages in hyperparameters, our DualTeacher is both effective and practical for incremental learning of object detectors.

\section{Conclusion}
In this work, we present semi-supervised incremental object detection (SSIOD) as a realistic setting for incremental learning of object detectors. In contrast to supervised incremental object detection, the coexistence of old and new classes in massive unlabelled data becomes a particular challenge, which severely affects the effectiveness of widely-used knowledge distillation strategies. To overcome this challenging issue, we leverage the property of non-overlapping annotations and the resulting exclusivity of predictions in incremental learning, concatenating as pseudo-labels the predictions of two teachers dedicated for old and new classes, respectively.
Extensive experiments demonstrate the validity and practicality of our approach.
We expect this work to provide a new perspective as well as a strong baseline for incremental learning of object detectors in real-world applications.
Subsequent work could further explore the effective usage of limited supervision and massive unlabelled data of dynamic distributions.

\section*{Acknowledgements}
This work was supported by the National Key Research and Development Program of China (No. 2020AAA0106302), NSFC Projects (Nos.~62061136001, 92248303, 62106123, 61972224), BNRist (BNR2022RC01006), Tsinghua Institute for Guo Qiang, and the High Performance Computing Center, Tsinghua University. L.W. is also supported by Shuimu Tsinghua Scholar, and J.Z. is also supported by the XPlorer Prize.

\bibliography{aaai24}

\clearpage
\appendix
\section{Implementation Details}
We summarize the implementation details as below.

\textbf{Task split.} The MS-COCO dataset contains object instances of 80 classes. We consider three different task splits, including (1) five subsets of 16 classes each, i.e., 5-phase; (2) two subsets of 40 classes each, i.e., 2-phase, 40+40 classes; and (3) two subsets of 60 classes and 20 classes, i.e., 2-phase, 60+20 classes.

\textbf{Implementation.} Our work is implemented in the framework of Unbiased Teacher \cite{liu2021unbiased}, which is a representative semi-supervised learning technique for object detection. Specifically, we use a similar Faster R-CNN architecture with ResNet-50 as the backbone, initialized by an ImageNet pre-trained model. We use random horizontal flip for weak augmentation, and use color jittering, grayscale, Gaussian blur, and cutout patches for strong augmentation. We employ a SGD optimizer with learning rate of 0.01 and momentum of 0.9. The batch size for both labelled and unlabelled data is set to 16. We first execute a burn-in stage for 2k iterations, followed by the teacher-student mutual learning stage. Following the implementation of the Unbiased Teacher, we trained all methods for a total of 180k iterations, distributing the number of iterations according to the size of the training dataset for each task. 


\textbf{Hyperparameters.} The hyperparameters for semi-supervised learning follow the official implementation of Unbiased Teacher, summarized in Table~\ref{table_hyper_parameters}.
In particular, we filter the pseudo-labels of the new and old teachers based on the network scores, retaining only the pseudo-labels with network scores greater than confidence threshold $\delta = 0.7$ to train the student model. 
Note that our approach does not introduce additional hyperparameters on the top of semi-supervised learning, thus avoiding the procedure of hyperparameter search.

\textbf{Environment.} For all experiments, we use 4 Tesla V100-SXM2-32GB GPUs with the CentOS 7 operating system, CUDA version 9.2, PyTorch version 1.7.0, and Detectron2 version 0.5. Due to the large computational overhead of semi-supervised learning, the reported performance is obtained from running once the corresponding task sequence.

\begin{table}[ht]
	\centering
    \caption{Summary of hyperparameters in experiments. $^*$The distillation loss is used for ILOD / Faster ILOD rather than our approach.} 
      \vspace{-0.1cm}
	\smallskip
      \renewcommand\arraystretch{1.4}
    \small{
	\resizebox{0.48\textwidth}{!}{ 
	\begin{tabular}{llc}
	 \hline
        Hyperparameter & Description & Value \\
        \hline
        $\delta$ & Conﬁdence threshold & 0.7 \\
        $bs_{{\rm{sup}}}$ & Batch size for labelled data & 16 \\
        $bs_{{\rm{unsup}}}$ & Batch size for unlabelled data & 16 \\
        $\alpha_{{\rm{tea}}}$ & EMA rate & 0.9996 \\
        $\alpha_{{\rm{stu}}}$ & Learning rate & 0.01 \\
        $\lambda_{{\rm{unsup}}}$ & Strength of unsupervised loss & 4 \\
        $^* \lambda_{{\rm{fea}}}^{{\rm{kd}}}$ & \makecell[l]{Strength of feature distillation loss} & 1.0 \\ 
        $^* \lambda_{{\rm{rpn}}}^{{\rm{kd}}}$ & \makecell[l]{Strength of RPN distillation loss} & 0.001 \\ 
        $^* \lambda_{{\rm{roi}}}^{{\rm{kd}}}$ & \makecell[l]{Strength of ROI distillation loss} & 0.1 \\ 
       \hline
	\end{tabular}
	} }
	\label{table_hyper_parameters}
	\vspace{-0.2cm}
\end{table}

\section{Extended Results}

\begin{figure}[th]
    \centering
    \includegraphics[width=0.85\linewidth]{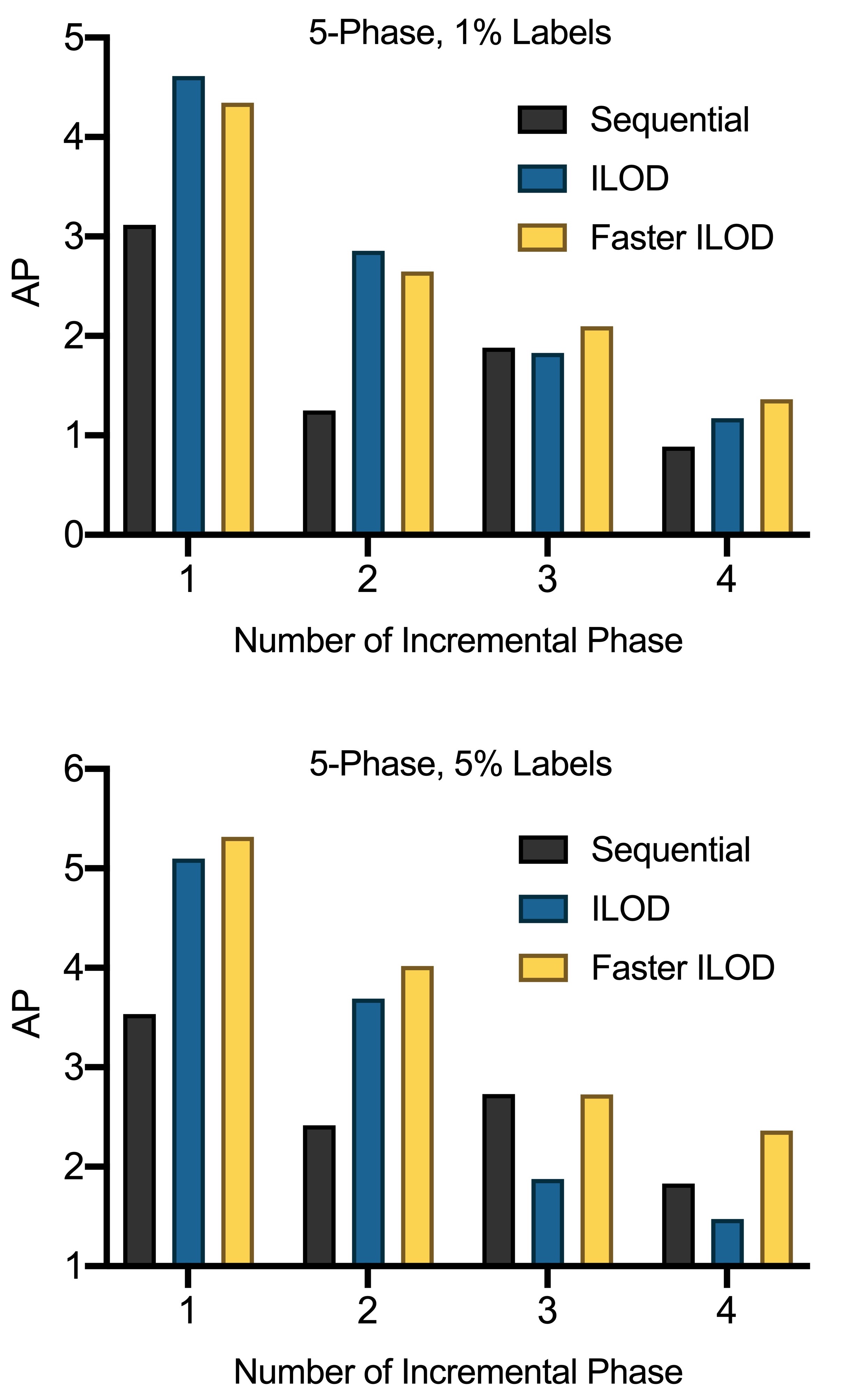}
\caption{Performance of supervised IOD. Here we compare sequential training and Faster ILOD under the 5-phase setting of only 1\% and 5\% labelled data (i.e., no unlabelled data).} 
\vspace{-0.2cm}
\label{supervised_IOD}
\end{figure}

\begin{figure*}[th]
\hsize=\textwidth
    \centering
    \includegraphics[width=0.98\linewidth]{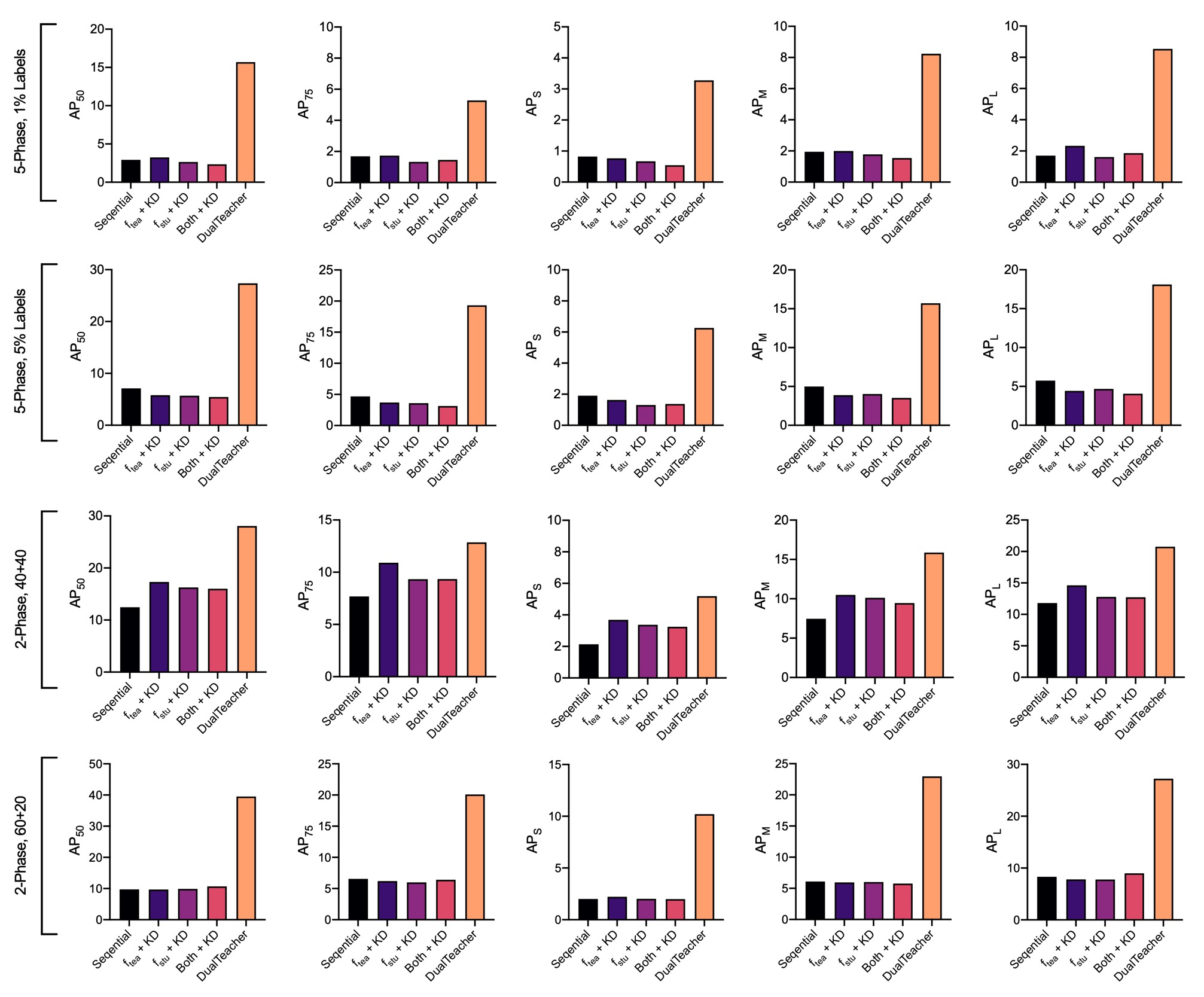}
\caption{Extended results of SSIOD with different metrics of average precision. Here we take Faster ILOD as the baseline approach of knowledge distillation (KD), which can be implemented to $f_{{\rm{tea}}}$, $f_{{\rm{stu}}}$, or both.} 
\vspace{-0.2cm}
\label{detection_ap_all}
\end{figure*}

We present two extended results as below. The first is the performance of supervised IOD (Fig.~\ref{supervised_IOD}). We use only 1\% and 5\% labelled data for incremental learning, where representative methods of knowledge distillation achieve strong improvements over sequential training. 
Therefore, such methods fail in SSIOD specifically because of the large amount of unlabelled data.
The second is the performance of SSIOD with different metrics of average precision (Fig.~\ref{detection_ap_all}), which more comprehensively demonstrates the superiority of our DualTeacher compared to knowledge distillation. 

\end{document}